\documentclass[conference]{IEEEtran}
\IEEEoverridecommandlockouts

\usepackage{cite}
\usepackage{amsmath,amssymb,amsfonts}
\usepackage{algorithmic}
\usepackage{graphicx}
\usepackage{textcomp}
\usepackage{xcolor}
\usepackage[acronym,toc,nonumberlist]{glossaries}
\usepackage{tikz}
\usetikzlibrary{matrix,calc}
\usepackage{pgfplots}

\def\BibTeX{{\rm B\kern-.05em{\sc i\kern-.025em b}\kern-.08em
    T\kern-.1667em\lower.7ex\hbox{E}\kern-.125emX}}

\begin{document}

\title{Fuzzy Fingerprinting Transformer Language-Models for Emotion Recognition in Conversations\\

\thanks{This work was supported by Fundação para a Ciência e a Tecnologia (FCT), through Portuguese national funds Ref. UIDB/50021/2020, Agência Nacional de Inovação (ANI), through the project CMU-PT MAIA Ref. 045909, RRP and Next Generation EU project Center for Responsible AI Ref. C645008882-00000055, and the COST Action Multi3Generation Ref. CA18231.}
}

\author{\IEEEauthorblockN{Patrícia Pereira}
\IEEEauthorblockA{\textit{INESC-ID} \\
\textit{Instituto Superior Técnico}\\
\textit{Universidade de Lisboa}\\
Lisbon, Portugal \\
{\tt \scriptsize ~~~~~~~~~~patriciaspereira@tecnico.ulisboa.pt~~~~~~~~~~}}
\and
\IEEEauthorblockN{Rui Ribeiro}
\IEEEauthorblockA{\textit{INESC-ID} \\
\textit{Instituto Superior Técnico}\\
\textit{Universidade de Lisboa}\\
Lisbon, Portugal \\
{\tt \scriptsize~~~~~~~~~~~~~~~~~rui.m.ribeiro@tecnico.ulisboa.pt~~~~~~~~~~~~~~~~~}}
\and
\IEEEauthorblockN{Helena Moniz}
\IEEEauthorblockA{\textit{INESC-ID} \\
\textit{Faculdade de Letras}\\
\textit{Universidade de Lisboa}\\
Lisbon, Portugal \\
{\tt \scriptsize ~~~~~~~helena.moniz@inesc-id.pt~~~~~~~}}

\and
\IEEEauthorblockN{Luisa Coheur}
\IEEEauthorblockA{\textit{INESC-ID} \\
\textit{Instituto Superior Técnico}\\
\textit{Universidade de Lisboa}\\
Lisbon, Portugal \\
{\tt \scriptsize luisa.coheur@tecnico.ulisboa.pt}}
\and
\IEEEauthorblockN{Joao Paulo Carvalho}
\IEEEauthorblockA{\textit{INESC-ID} \\
\textit{Instituto Superior Técnico}\\
\textit{Universidade de Lisboa}\\
Lisbon, Portugal \\
{\tt \scriptsize ~~~~joao.carvalho@inesc-id.pt~~~~}}
}

\maketitle

\begin{abstract}
Fuzzy Fingerprints have been successfully used as an interpretable text classification technique, but, like most other techniques, have been largely surpassed in performance by Large Pre-trained Language Models, such as BERT or RoBERTa. These models deliver state-of-the-art results in several Natural Language Processing tasks, namely Emotion Recognition in Conversations (ERC), but suffer from the lack of interpretability and explainability. In this paper, we propose to combine the two approaches to perform ERC, as a means to obtain simpler and more interpretable Large Language Models-based classifiers. We propose to feed the utterances and their previous conversational turns to a pre-trained RoBERTa, obtaining contextual embedding utterance representations, that are then supplied to an adapted Fuzzy Fingerprint classification module.
We validate our approach on the widely used DailyDialog ERC benchmark dataset, in which we obtain  state-of-the-art level results using a much lighter model.

\end{abstract}

\begin{IEEEkeywords}
Fuzzy Fingerprints, Large Language models, RoBERTa, Emotion Recognition in Conversations
\end{IEEEkeywords}

\section{Introduction}
\label{introduction}

Emotion Recognition in Conversations (ERC) has become increasingly important with the widespread use of conversational agents.
Recognizing emotions is essential for communication, being a crucial component in the development of effective and empathetic conversational agents.
ERC is also used for automatic opinion mining and therapeutic practices.

\begin{figure}[!ht]
\begin{center}
  \includegraphics[width=0.9\linewidth]{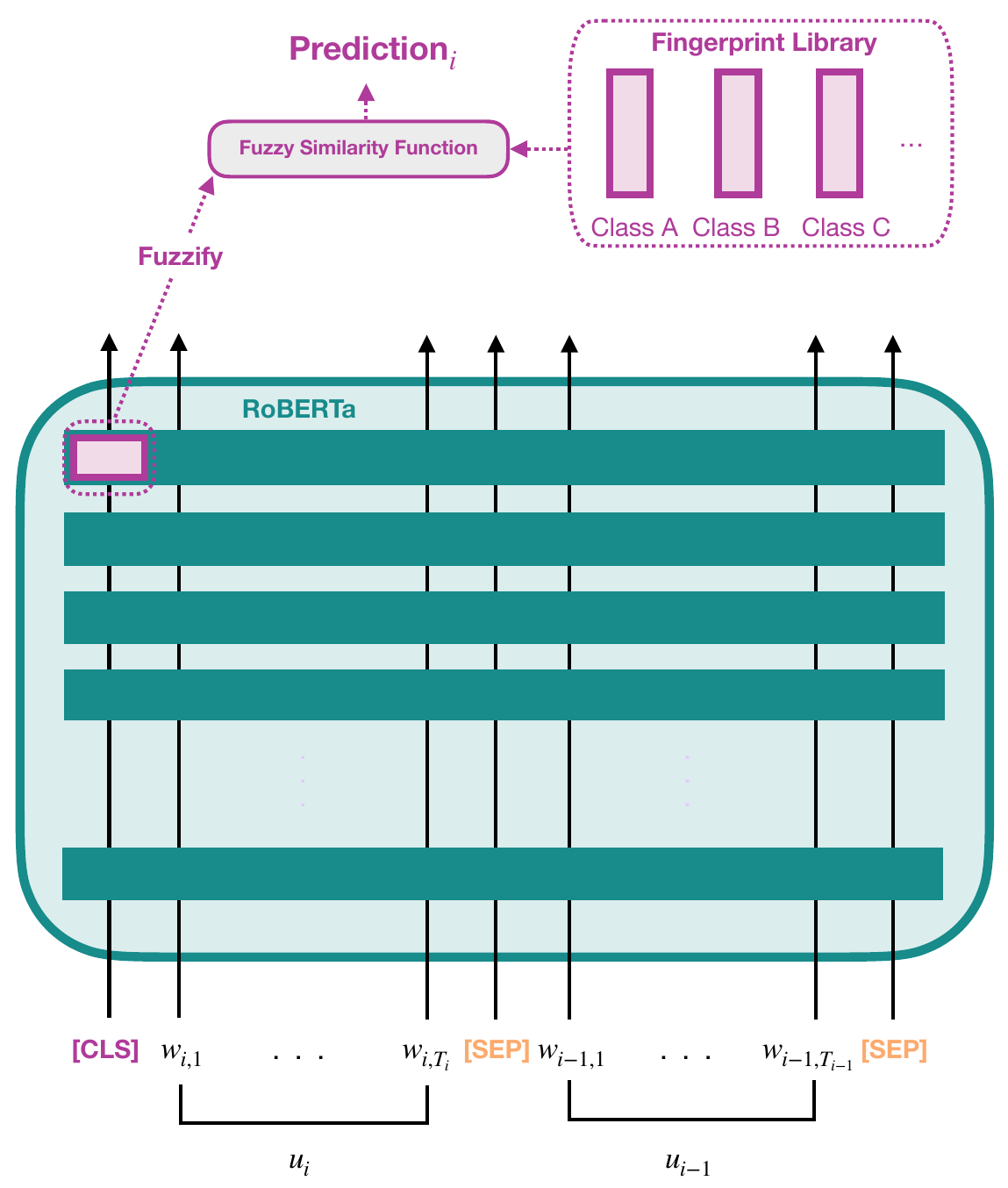}
  
 \end{center}
  \caption{Model architecture. In this example input, an utterance, $u_{i}$, and its conversational context, $u_{i-1}$, are fed to RoBERTa encoder, of which the \texttt{[CLS]} token of the last layer goes through the Fuzzy Fingerprint module.
}
  \label{f1}
  \end{figure}

There is thus a growing interest in endowing machines with efficient emotion recognition modules.

Fuzzy Fingerprints (FFP) have been successfully used as an interpretable text classification technique \cite{homem2011authorship}\cite{rosa2018using}, but, like most other text classification techniques, have been largely surpassed in performance by Pre-trained Language-Models (PLMs), which are the current state-of-the-art in most Text Classification tasks, including ERC. 
However, PLMs are massively complex black boxes that lack interpretability, and explaining the obtained outcomes from these models is still a challenge. Like other deep neural-based models, despite being able of state-of-the-art performance, they can also fail catastrophically in situations of apparent simplicity \cite{goodfellow2014explaining}.

In this paper, we propose to combine our ERC context-dependent embedding representation from the RoBERTa PLM \cite{pereira2023context} with an adapted Fuzzy Fingerprint classification module \cite{RIBEIRO-FUZZYFINGERPRINTING} in order to introduce some interpretability to RoBERTa, while maintaining PLMs performance levels.
To test the proposed approach we resort to Daily Dialog \cite{li2017dailydialog}, a widely used benchmark ERC dataset, and we find that the addition of FFP not only allows for result interpretability but also results in a performance increase that gives a state-of-the-art comparable performance with much heavier models.

\section{Related Work}
\label{related}

\subsection{Emotion Recognition in Conversations}

Knowledge and understanding of the conversational context, i.e., the previous conversational turns, are extremely valuable for identifying the emotions of the interlocutors \cite{poria2019emotion}\cite{pereira2022deep}. Therefore, in this section, we describe approaches that leverage the conversational context in ERC.

Amongst the first works considering contextual interdependences among utterances is the one by Poria et al. \cite{poria2017context}, which leverages Long Short-Term Memory networks (LSTMs) to extract contextual features from the utterances.
A more elaborate model is DialogueRNN \cite{majumder2019dialoguernn}, which uses three Gated Recurring  Units (GRU) in the classification module to model several aspects of the conversation.

One major problem in using RNNs is the long path of information flow, that difficults the capture of long-term dependencies. 
These dependencies can be better captured with the Transformer architecture which has a shorter path of information flow. Its introduction in 2017 \cite{vaswani2017attention} led to a new state-of-the-art in several Natural Language Processing (NLP) tasks. 
Amongst the first works leveraging the Transformer is the Knowledge-Enriched Transformer (KET) \cite{zhong2019knowledge}.

Following the emergence of Transformers, new encoder PLMs based on this architecture (such as BERT \cite{devlin2018bert} and RoBERTa \cite{liu2019roberta}) were introduced and achieved state-of-the-art in various NLP benchmarks. Since their emergence, most state-of-the-art ERC works resorted to encoder PLMs.

COSMIC \cite{ghosal2020cosmic} leverages RoBERTa to encode utterances. Furthermore, it makes use of the commonsense transformer model COMET \cite{bosselut2019comet}, a large knowledge base, in order to extract commonsense features. Five bi-directional GRUs model several aspects of the conversation in the classification module.
Psychological \cite{li2021past} also uses RoBERTa for utterance encoding and the large external knowledge base COMET. It introduces a graph of utterances for conversation-level encoding that needs to be processed with a graph transformer.

Contrarily to the aforementioned approaches, we produce RoBERTa context-dependent embedding representations of each utterance, discarding the need for such complex classification modules. 

\subsection{Pre-trained Transformer Encoder
Language-Models} 

A very popular large pre-trained model is BERT \cite{devlin2018bert}, a multi-layer bidirectional Transformer encoder trained to perform language modeling and next-sentence prediction.
BERT was trained on English Wikipedia and the BooksCorpus during a computationally expensive process, where it learns deep contextual embeddings, i.e, vectors representing the semantics of each word or sequence of words.

RoBERTa \cite{liu2019roberta}, a successor to BERT, contains the same architecture as BERT but is pre-trained with more data and for a longer period of time, uses larger mini-batches and a larger learning rate, and discards BERT's task of next-sentence prediction.

RoBERTa has outperformed BERT in various tasks using the same amount of data.

PLMs are pre-trained on certain NLP tasks, and in order to be adapted to specific tasks, they just need to be fine-tuned to the task at hand. The process of fine-tuning consists of supervised training of the PLM in the target dataset, after appending a classification module to fit the dataset characteristics, namely the set of classification labels. During this process, the PLM's weights adjust to deliver maximal performance in the target dataset.

\subsection{Fuzzy Fingerprints}

Fingerprint identification is a well-known and widely documented technique in forensic sciences. In Computer Science, a fingerprint is a procedure that maps an arbitrarily large data item to a much more compact information block (a fingerprint) that uniquely identifies the original data for all practical purposes.

Fuzzy Fingerprints (FFP), as used in this work, were introduced as a technique to identify one-out-of-many-suspects in tasks such as Web User Identification \cite{homem2011web} or Text Authorship Identification \cite{homem2011authorship}. For this purpose, FFP are built based on feature frequency. For example, for text classification purposes, we consider a set of texts associated with a given class to build the class fingerprint and can use the frequency of each word in each text to build the fingerprint for that class. 

Fuzzy Fingerprints have been further used and adapted for several tasks involving text classification, such as Tweet Topic Detection \cite{rosa2014twitter} or cyberbullying detection in social networks \cite{rosa2018using}. 

\subsubsection{Fuzzy Fingerprint Creation and Fuzzy Fingerprint Libraries}
\label{sec:ffp-creation}
The training set is processed to compute the top-$K$ feature list for each class. Consider $F_j$ as the set of events of class $j$ (simplistic example: the set of all words for all texts belonging to class $j$). The result consists of a list of $K$ pairs $\{v_i, n_i\}$, $1<i\le K$, where $v_i$ is the $i$-th most frequent feature and $n_i$ the corresponding count. 
The next step consists in fuzzifying each top-$K$ list: a membership value is assigned to each feature in the set based on the order in the list (the rank). 
The more frequent features will have a higher membership value. 
The FFP ($\Phi$), consists of a size-$K$ fuzzy vector where each position $i$ contains element $v_i$ and a membership value $\mu_i$. 
A class $j$ will be represented by its fingerprint $\Phi_j = \Phi(F_j)$. Formally, a fingerprint  $\Phi_j  = \{( v_{ji}, \mu_{ji}) \mid i = 1 \ldots k_j\}$ has length $k_j$, with $S_j = \{v_{ji} \mid i = 1 \ldots k_j\}$ representing the set of $v$’s in $\Phi_j$. The set of all class fingerprints will constitute the fingerprint library.

\subsubsection{Fuzzy Fingerprint Detection}

In order to find the class of an unknown instance, for example, a text $T$, we start by computing the size-$K$ fingerprint of $T$, $\Phi_T$. Then we compare the fingerprint of $T$ with the fingerprints $\Phi_j$ of all classes present in the fingerprint library. The unknown text is classified as $j$ if it has the most similar fingerprint to $\Phi_j$. Fingerprint similarity, $\mbox{sim}(\Phi_T, \Phi_j)$, is calculated using
\begin{equation}
\mbox{sim}(\Phi_T, \Phi_j) = \sum_{v \in S_T \cup S_j} \frac{\min(\mu_v(\Phi_T), \mu_v(\Phi_j))}{N},
\label{eq:FFFcomparison}
\end{equation}
where $\mu_v(\Phi_x)$ is the membership value associated with the rank of element $v$ in fingerprint $x$. This function is based on the fuzzy AND (here we use the minimum or Gōdel t-norm). $N$ is a constant that can be used for normalization purposes.

\section{Fuzzy fingerprinting RoBERTa for ERC}
\label{methods}
\subsection{Task Definition}

Given a conversation composed of a sequence of $u_{i}$ utterances with corresponding $emotion_{i}$ from a predefined set of emotions, the aim of ERC is to correctly assign an emotion to each utterance of the conversation.  An utterance consists of a sequence of $w_{it}$ tokens representing its $T_{i}$ words:

\begin{equation}\label{equ}
u_{i}=(w_{i1},w_{i2},...,w_{iT_{i}}).
\end{equation}

\subsection{Context-Dependent Embedding Utterance Representations}

The most common approach for ERC has been to produce context-independent representations of each utterance (using PLMs), and subsequently perform contextual modeling of the obtained representations with classification modules comprising gated and graph neural networks. 
In a recent work \cite{pereira2023context}, we proposed to produce context-dependent representations of each utterance that represent not only the utterance but also a given number of previous utterances from the conversation. This context-based approach allowed us to discard the need for complex classification modules: a single fully connected linear softmax layer appended to this variation of RoBERTa, is enough to achieve state-of-the-art level performance.

When pooling the embeddings, we chose the first embedding from the last layer L, the \texttt{[CLS]} token which is used for classification, as in Equation \ref{ecls}. 
\begin{equation}\label{ecls}
pooled_{i}=RoBERTa_{L,[CLS]}(input_{i}).
\end{equation}

This embedding is then fed to a fully connected linear layer with softmax so that the complete model maximizes the probability of the correct labels. 

For this work, we use the fully connected layer to fine-tune RoBERTa, but discard and replace it afterward with a Fuzzy Fingerprint classification module, as detailed in the next section and represented in Figure \ref{f1}.

\subsection{Fuzzy Fingerprinting RoBERTa}

Fuzzy Fingerprints are based on the concept of feature frequencies (section \ref{sec:ffp-creation}). In order to fingerprint RoBERTa, we must find a way to adapt this concept to   
 the RoBERTa's output, i.e., to the final hidden state of the \texttt{[CLS]} classification token, which  is a real-valued vector (with dimension 768) where each element does not have a known meaning.
 
 A solution to address this issue, is to use the intensity of the activation of each element from the RoBERTa's output as a proxy for feature frequency. 

Unlike traditional FFP, where the size of the universe of discourse is finite but unknown (e.g. the number of all existing words), here the fingerprint size is limited to 768. Hence it is possible to define a RoBERTa FFP as a discrete Fuzzy Set (in the discrete universe of the RoBERTa outputs), where each of the 768 outputs has an associated membership value that is computed based on its activation rank. Only the top-$K$ elements have a membership greater than zero, and for practical purposes, the set is ordered by the membership function of its elements.   
 
 The process to create the  fingerprint of a specific emotion can be therefore succinctly described as ``ranking and fuzzifying the activation of the RoBERTa's outputs through the training set (of that emotion)''. The procedure is described as follows:
\begin{enumerate}
    \item The training data is used to create fuzzy fingerprints for each emotion (class). 
    \item The fingerprint for each emotion begins as a 2D vector of size 768, where each position contains an index and a value initialized to $0$.
    \item The fine-tuned context-based RoBERTa is fed with all the training examples of a given emotion (one by one).  
    \item The RoBERTa output for each example consists of a 768-sized vector of real values. The real-valued outputs are added to the fingerprint. Hence, after all the training examples (of a given emotion) are fed into RoBERTa, the fingerprint of the emotion consists of the 2D vector, where each position contains an index and the accumulated value of the RoBERTa output for all the examples.
    \item Order the fingerprint vector by the accumulated value (in descending order).
    \item Reduce the 2D vector to a single dimension by discarding the column containing the accumulated values (only the rank matters). As a result, we have, for each emotion, a vector of dimension 768, containing the indexes of the mostly activated RoBERTa outputs for that emotion, i.e., the RoBERTa outputs are ranked by activation on the training set.
    \item The fingerprint will only use the top-$K$ RoBERTa outputs for classification purposes (instead of the whole 768 RoBERTa outputs). $K$ is selected on the validation set.
    \item The FFP is obtained by fuzzifying the top-$K$ sized vector according to the following function:

\begin{equation}\label{mu}
\mu_{i}=1-\frac{a \times i}{K}, \forall a \in [0,1]
\end{equation}

in which $i$ is the index of the element in the sorted vector, $K$, is the fingerprint size, and $a$ adjusts the slope of the function.  The function gives higher-ranked items a larger membership. Other functions were tested on a validation set. This is the function that provided the best score.

\end{enumerate}

After obtaining the fingerprints for all possible emotions (the Fingerprint Library), classification can be performed. Given a sequence of $u_{i}$ utterances to be classified:
\begin{enumerate}
    \item Pass $u_{i}$ through RoBERTa.
    \item Create the fingerprint of $u_{i}$ using the same procedure used to create the fingerprint of an emotion (i.e. rank the activation of the output vector, select the Top-K elements and fuzzify the resulting vector).
    \item Check the similarity of the fingerprint of $u_{i}$ against the fingerprint of each emotion using the Fuzzy Fingerprint similarity function from Equation \ref{eq:FFFcomparison}, and select the emotion with the highest similarity.
\end{enumerate}

\section{Experimental Setup}
\label{setup}

\subsection{Training Details}

To obtain the context-dependent embedding utterance representations we leverage RoBERTa-base from the Transformers library by Hugging Face \cite{wolf2019huggingface}, trained with the cross-entropy loss with logits. We use the Adam optimizer, with an initial learning rate of 1e-5 and 5e-5, for the encoder and the classification head, respectively with a layer-wise decay rate of 0.95 after each training epoch for the encoder, which is frozen for the first epoch. The batch size is set to 4. Gradient clipping is set to 1.0. Early stopping is used to terminate training if there is no improvement after 5 consecutive epochs on the validation set over macro-F1, for a maximum of 10 epochs. The checkpoint used in testing is the one that achieves the highest macro-F1 score on the validation set. The Fuzzy Fingerprint module was proposed and coded by us.

From Table \ref{t1} it can be observed the DailyDialog dataset is imbalanced, not only for its dominant majority neutral class but also for the relative imbalance between minority classes. To promote consistent performance across all classes we have opted to use the macro-F1 score for model selection. 

With regards to the parameter $a$ we have observed experimentally that 0.8 is a suitable value.

\subsection{Evaluation}

The reported results are an average of 5 runs corresponding to 5 distinct random seeds that are kept for a meaningful comparison of all experiments. This average is motivated by the fact that results for the same experiment obtained with different random seeds can have high variability in the macro F1-score, comparable to the improvements that we report regarding state-of-the-art models. This procedure is also in line with several authors that also resort to 5 run averages in ERC tasks \cite{li2021past} \cite{zhong2019knowledge} \cite{shen2020dialogxl} \cite{shen2021directed}.

\subsection{Dataset}

DailyDialog \cite{li2017dailydialog} is built from websites used to practice English dialogue in daily life. Table \ref{t0} resumes its main statistics. It is labeled with the six Ekman’s basic emotions \cite{ekman1999basic}, anger, disgust, fear, happiness, sadness and surprise, or neutral. The publicly available splits of Yanran are used and the label distribution is presented in Table \ref{t1}.

\begin{table}[htbp]
  \caption{Statistics of the DailyDialog dataset}
 \centering
  \begin{tabular}{lcc}
    \hline 
    Num. dialogues &13,118\\
    Num. turns/labels &102,979\\
     Avg. turns per dialogue &7.9\\

    \hline
  \end{tabular} 
 \label{t0}
\end{table}

\begin{table}[htbp]
  \caption{Proportion of labels in the DailyDialog dataset}
 \centering
  \begin{tabular}{cccccccc}
    \hline
     
  \textbf{Ang} & \textbf{Disg} & \textbf{Fear} & \textbf{Hap} & \textbf{Sad} & \textbf{Sur} & \textbf{Neu} \\
   
   1.0\%&0.3\%& 0.2\%& 12.5\% &1.1\%& 1.8\%&83.1\% \\

    \hline
  \end{tabular} 
 \label{t1}
\end{table}

\section{Results and Analysis}
\label{results}

In this section, we present some experimental results and compare our approach to current state-of-the-art models.

\subsection{Fuzzy Fingerprint Size - $K$}

 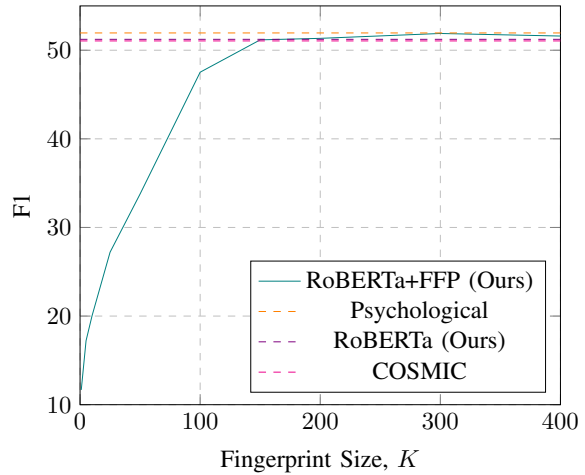
\begin{figure}[htpb]
 \centering
    \resizebox{0.9\columnwidth}{!}{
        \begin{tikzpicture}
        \begin{axis}[
            title={},
            xlabel={Fingerprint Size, $K$},
            ylabel={F1},
            xmin=0, xmax=400,
            ymin=10, ymax=55,
            legend pos=north west,
            ymajorgrids=true,
            grid style=dashed,
            grid=both,
            legend pos=south east,
            ylabel near ticks, yticklabel pos=left,
        ]
        
        \addplot[color=teal, mark=.] table[x=K, y=F1, col sep=semicolon, /pgf/number format/read comma as period] {plot.csv}; 
        \addlegendentry{RoBERTa+FFP (Ours)}

        \addplot[mark=none, orange, dashed, samples=2, domain = 0:400] {51.95};
        \addlegendentry{Psychological}

        \addplot[mark=none, violet, dashed, samples=2, domain = 0:400] {51.23};
        \addlegendentry{RoBERTa (Ours)}

        \addplot[mark=none, magenta, dashed, samples=2, domain = 0:400] {51.05};
        \addlegendentry{COSMIC}

        \end{axis}
        \end{tikzpicture}
    }
\caption{Variation of the F1-score with the fingerprint size $K$. Other models use all RoBERTa outputs, hence having an equivalent $K$=768}  

 \label{kf1}
\end{figure}

We start by checking the effect of the FFP size $K$ on the performance. 
From Table \ref{tkf1} and the graph in Figure \ref{kf1} we can see that, for $K$ larger than 150, the performance of our proposed model is comparable to using all the 768 RoBERTa outputs (and within state-of-the-art performance level). It is therefore possible to conclude that it is not necessary to use all the RoBERTa outputs to obtain high performance and that it is possible to use a smaller and less computationally demanding model once $K$ is decided. This also hints that it is possible to train a smaller base model for this task.

\begin{table}[!htpb]
 \centering
 \caption{Variation of the F1-score with the fingerprint size $K$}
  \begin{tabular}{cccccccccccc}
    \hline
       
       \textbf{K}& 1&5&10&25&50\\
       \textbf{F1}& 11.67&17.22&20.04&27.21&33.78\\
           \hline
        \textbf{K}&100&150&200&300&400\\
       \textbf{F1}&47.52&51.17&51.34&\textbf{51.89}&51.60 \\
    \hline
    \multicolumn{5}{l}{\textbf{F1 without the Fingerprints module:}} &51.23\\

    \hline
  \end{tabular} 
 \label{tkf1}
\end{table} 

It can be observed that the peak of performance happens for $K$=300, which yields an F1-score of 51.89.

\subsection{Comparison without the Fuzzy Fingerprint module}

When using the same RoBERTa representation for the Fuzzy Fingerprint module but resorting only to the simple fully connected linear layer as a classification module both for training and prediction, we obtain an F1-score of 51.23 \cite{pereira2023context}. Given that performance differences from state-of-the-art approaches can consist in improvements of the magnitude of less than 1 F1-score, as it can be seen in Table \ref{t5}, our improvement of 0.66 in F1-score is considered significant (results are an average of 5 runs to preserve statistical significance).

\subsection{Performance on each emotion label}

We report the F1-score on each individual emotion label with the best value for $K$=300 in Table \ref{t4} (5 runs average). 

\begin{table}[htpb]
 \centering
 \caption{Model performance on each individual emotion label}
  \begin{tabular}{cccccccc}
    \hline
     &\textbf{Ang} & \textbf{Disg} & \textbf{Fear} & \textbf{Hap} & \textbf{Sad} & \textbf{Sur} & \textbf{Neu} \\
    \hline
    \textbf{F1}& 43.22 &32.89  & 42.12& 61.38 &39.45& 52.89 &  91.30   \\

    \hline
  \end{tabular} 
 \label{t4}
\end{table}

From Table \ref{t4} it can be observed that the classifier performs better at the most represented classes in the dataset, having an F1 score of above 60 for the well-represented class Happiness and an even higher F1 score of around 90 for the over-represented Neutral class.

\subsection{Comparison with state-of-the-art}

We compare our approach to other state-of-the-art works that also resort to RoBERTa. This allows for a fair comparison between approaches given that using this PLM brings great performance increases when compared to using other means of utterance feature extraction. 

We compare our approach to COSMIC \cite{ghosal2020cosmic}, RoBERTa and RoBERTa DialogueRNN, implemented by the authors of COSMIC, and the Psychological model \cite{li2021past}, all models described in Section \ref{related}. Results are displayed in table \ref{t5} and are an average of 5 runs.

\begin{table}[!htpb]
 \centering
 \caption{Comparison with state-of-the-art works}
  \begin{tabular}{lccc}
    \hline
       
       &macro-F1  \\
    \hline
    RoBERTa \cite{ghosal2020cosmic} & 48.20\\
    RoBERTa + DialogueRNN \cite{ghosal2020cosmic} &49.65\\
   
    COSMIC \cite{ghosal2020cosmic} &51.05\\  
    Contextual RoBERTa \cite{pereira2023context} &51.23\\
    Psychological \cite{li2021past}&\textbf{51.95}&\\
    \hline
    Contextual RoBERTa + FFP &51.89& \\

    \hline
  \end{tabular} 
 \label{t5}
\end{table} 

Our approach outperforms not only the simple RoBERTa, but also RoBERTa in a more elaborate gated neural network model such as DialogueRNN and COSMIC. The Psychological model has a slightly higher performance than ours, but it uses RoBERTa Large, needs a heavy commonsense knowledge base, COMET \cite{bosselut2019comet}, and a complex graph of utterances as a conversational level encoder that needs to be processed with a graph transformer. In comparison, our model uses RoBERTa base fine-tuned using context, and a minimalist fuzzy fingerprint classification model.  

We have also performed experiments with ChatGPT 3 and 4, with the introduction of a variable number of context turns in the prompt, and observed that both ChatGPT versions do not outperform our context-dependent embedding utterance representation approach \cite{pereira2023context}, which is outperformed by our proposed fuzzy fingerprints approach.

\subsection{FFP Interpretability/Explainability Examples}

We claim that using FFP can add some much-needed interpretability and explainability to PLMs. Here we present three examples that help to support our claim.  
We use small fingerprints for visualization and exemplification purposes. Table \ref{cf} presents the generated Emotion FFP for $K$=10.

\begin{table*}[ht]
\centering
\begin{tabular}{lcccccccccc}
\hline

$FFP_{Neu}=$&
        \{(217,1),&(644,0.9),&(541,0.8),&(718,0.7),&(401,0.6),&(330,0.5),&(426,0.4),&(78,0.3),&(580,0.2),&(114,0.1)\}\\

$FFP_{Ang}=$&
        \{(8,1),&(679,0.9),&(204,0.8),&(292,0.7),&(651,0.6),&(573,0.5),&(111,0.4),&(624,0.3),&(184,0.2),&(309,0.1)\}\\

$FFP_{Dis}=$&
        \{(588,1),&(573,0.9),&(27,0.8),&(154,0.7),&(331,0.6),&(67,0.5),&(561,0.4),&(5,0.3),&(503,0.2),&(446,0.1)\}\\
     
$FFP_{Fear}=$&
        \{(588,1),&(313,0.9),&(655,0.8),&(406,0.7),&(736,0.6),&(349,0.5),&(624,0.4),&(371,0.3),&(292,0.2),&(8,0.1)\}\\
      
$FFP_{Hap}=$&
        \{(588,1),&(585,0.9),&(388,0.8),&(600,0.7),&(767,0.6),&(319,0.5),&(741,0.4),&(561,0.3),&(473,0.2),&(139,0.1)\}\\

$FFP_{Sad}=$&
        \{(371,1),&(588,0.9),&(5,0.8),&(156,0.7),&(4,0.6),&(93,0.5),&(550,0.4),&(402,0.3),&(519,0.2),&(422,0.1)\}\\
 
$FFP_{Sur}=$&
        \{(691,1),&(588,0.9),&(97,0.8),&(573,0.7),&(530,0.6),&(535,0.5),&(654,0.4),&(384,0.3),&(366,0.2),&(613,0.1)\}\\

\hline
\end{tabular}
\caption{Class Fingerprints ($K$=10)}
\label{cf}
\end{table*}

The first noticeable aspect is how output 588 is present at the top of 5 out of the 7 classes. This can be potentially leveraged to improve ERC performance using a multistage classifier, since 588 is not present in the Neutral FFP, which is the majority class and where most misclassifications end up.  

In Table \ref{examples} three examples of utterance and class fingerprints, and the respective similarity results (considering $N=1$), are presented.

\begin{table*}[ht]
\centering
\begin{tabular}{lcccccccccc}
\hline

Text: & \multicolumn{10}{l}{Sorry , sir . It’s the sale price .} \\
$FFP_{Sample}=$&
        \{(\textbf{5},1),&(\textbf{371},0.9),&(\textbf{156},0.8),&(\textbf{550},0.7),&(\textbf{93},0.6),&(\textbf{4},0.5),&(232,0.4),&(424,0.3),&(\textbf{402},0.2),&(442,0.1)\}\\

Class Similarity:&$Neu=0$ &$Ang=0$& $Dis=0.3$& $Fear$& $Hap=0$ &$Sad=\textbf{1.3}$ &$Sur=0$\\
\hline
Text: & \multicolumn{10}{l}{You still have not given me those files I ’ Ve asked you for .}\\

$FFP_{Sample}=$&
        \{(\textbf{8},1),&(\textbf{679},0.9),&(\textbf{309},0.8),&(624,0.7),&(\textbf{292},0.6),&(76,0.5),&(134,0.4),&(204,0.3),&(560,0.2),&(459,0.1)\}\\

Class Similarity:&$Neu=0$ &$Ang=\textbf{1}$& $Dis=0$& $Fear=0.1$& $Hap=0$ &$Sad=0$ &$Sur=0$\\

\hline
Text: & \multicolumn{10}{l}{Don't forget to give me the files I've asked you for}\\
$FFP_{Sample}=$&
        \{(\textbf{330},1),&(\textbf{644},0.9),&(\textbf{541},0.8),&(\textbf{217},0.7),&(\textbf{114},0.6),&(\textbf{426},0.5),&(211,0.4),&(\textbf{718},0.3),&(\textbf{401},0.2),&(553,0.1)\}\\

Class Similarity:&$Neu=\textbf{1.4}$ &$Ang=0$& $Dis=0$& $Fear=0$& $Hap=0$ &$Sad=0$ &$Sur=0$\\

\hline
\end{tabular}
\caption{Classification examples ($K$=10)}
\label{examples}
\end{table*}

The first example shows that there are many elements in common between the utterance and the class FFP and there is no doubt that we have a correct classification.

The second and third examples show an interesting case of an incorrect classification and how the FFP can explain such error. The FFP of the utterance "You still have not given me those files I've asked you for" clearly indicates a strong similarity to the Angry emotion, with many important features in common between the fingerprints. Yet, according to the dataset annotation, the utterance is supposed to be Neutral. This misclassification cannot be explained by the FFP, however, it could certainly be argued by a native English speaker that the utterance has a definite negative meaning. Due to the negative meaning, the most present emotion is probably more Anger than Neutral, revealing either an annotation error or some hidden knowledge from the annotator that cannot be extracted from the utterance. A more polite and neutral way of expressing the same message would be, for example, "Don't forget to give me the files I've asked you for" (the third example on the table). Here we see that the FFP clearly indicates a Neutral emotion despite the sentence expressing the same message, albeit with a different tone.

\section{Conclusion and Future Work}
\label{conclusion}

In this paper, we proposed to combine pre-trained Transformer Language Models with Fuzzy Fingerprints. Concretely, we obtained context-dependent embedding utterance representations from the RoBERTa PLM that were then fed to an adapted FFP classification module.
We validated our approach on the DailyDialog dataset of ERC, for which our architecture clearly competes with state-of-the-art classifier architectures that resort to much more complex classification modules while adding some interpretability and explainability.

In future work, we plan to perform a more detailed study of the interpretability and explainability potential and explore model size reductions based on Fuzzy Fingerprints.

\bibliographystyle{plain}

\end{document}